\documentclass[sigconf, nonacm]{acmart}

\usepackage{booktabs}
\usepackage{caption}
\usepackage{graphicx}
\usepackage{multirow}
\usepackage{array}
\usepackage{makecell}
\usepackage{threeparttable}
\usepackage{wrapfig}
\usepackage{pifont}
\usepackage{fancyvrb}
\usepackage{fvextra}
\usepackage{floatflt}
\usepackage{comment}
\usepackage{tabularx}
\usepackage[ruled,vlined,linesnumbered]{algorithm2e}

\usepackage{hyperref}
\usepackage{url}
\usepackage{enumitem}
\setlist{topsep=0pt, leftmargin=*}
\usepackage{booktabs}
\usepackage{subfig}
\usepackage{natbib}

\usepackage{amsmath,amsfonts,bm}

\def\eqref#1{equation~\ref{#1}}

\def\1{\bm{1}}

\DeclareMathAlphabet{\mathsfit}{\encodingdefault}{\sfdefault}{m}{sl}
\SetMathAlphabet{\mathsfit}{bold}{\encodingdefault}{\sfdefault}{bx}{n}

\newcommand{\sys}{\textsc{AgentSM}}

\newcommand{\heading}[1] {{\hfill\break\noindent{\textbf{\emph{#1.}}} }}

\begin{document}
\title{\sys{}: Semantic Memory for Agentic Text-to-SQL}

\author{Asim Biswal$^{\triangle\circ}$, Chuan Lei$^{*\diamond}$, Xiao Qin$^{*\Box}$, Aodong Li$^{\circ}$, \\ Balakrishnan Narayanaswamy$^{\circ}$, Tim Kraska$^{\circ}$}
\affiliation{
  \institution{\textsuperscript{$\circ$}\textit{Amazon Web Services} 
  \quad\textsuperscript{$\triangle$}\textit{University of California, Berkeley} 
  \quad\textsuperscript{$\diamond$}\textit{Oracle Corporation} 
  \quad \textsuperscript{$\Box$}\textit{Snowflake Inc.}}
}
\email{abiswal@berkeley.edu,chuan.lei@oracle.com,xiao.qin@snowflake.com}
\email{{aodongli,muralibn,timkrask}@amazon.com}
\thanks{*Work done at Amazon Web Services.}

\begin{abstract}
Recent advances in LLM-based Text-to-SQL have achieved remarkable gains on public benchmarks such as BIRD and Spider. Yet, these systems struggle to scale in realistic enterprise settings with large, complex schemas, diverse SQL dialects, and expensive multi-step reasoning. Emerging agentic approaches show potential for adaptive reasoning but often suffer from inefficiency and instability—repeating interactions with databases, producing inconsistent outputs, and occasionally failing to generate valid answers. To address these challenges, we introduce Agent Semantic Memory (\sys{}), an agentic framework for Text-to-SQL that builds and leverages interpretable semantic memory. Instead of relying on raw scratchpads or vector retrieval, \sys{} captures prior execution traces—or synthesizes curated ones—as structured programs that directly guide future reasoning. This design enables systematic reuse of reasoning paths, which allows agents to scale to larger schemas, more complex questions, and longer trajectories efficiently and reliably. Compared to state-of-the-art systems, \sys{} achieves higher efficiency by reducing average token usage and trajectory length by 25\% and 35\%, respectively, on the Spider 2.0 benchmark. It also improves execution accuracy, reaching a state-of-the-art accuracy of 44.8\% on the Spider 2.0 Lite benchmark.
\end{abstract}

\maketitle



\section{Introduction}
\label{sec:intro}

Text-to-SQL seeks to enable interaction with structured data by translating natural language tasks into executable SQL queries. This capability is particularly valuable in enterprise data analytics and business intelligence, where non-technical users need to extract insights from large, complex databases without mastering SQL or detailed schema knowledge.

Recent advances in large language models (LLMs)~\cite{Zorpette_2025}, prompting strategies~\cite{chasesql}, and post-training techniques~\cite{arctictext2sql} have driven notable progress in Text-to-SQL performance across benchmarks such as BIRD~\citep{li2024can} and Spider~\citep{yu2019spiderlargescalehumanlabeleddataset}. However, most existing Text-to-SQL systems remain difficult to scale in realistic enterprise settings, where challenges such as deep nested schemas, diverse SQL dialects, and domain-specific business logic lead to degraded accuracy and efficiency.

To evaluate Text-to-SQL systems under more realistic conditions, the Spider 2.0 benchmark~\citep{lei2024spider} was recently introduced, featuring complex, multi-dialect databases and extremely long contexts. Traditional Text-to-SQL approaches—including vector-based schema retrieval~\cite{10.14778/3749646.3749723}, candidate generation with majority voting~\cite{chasesql}, and self-consistency decoding~\cite{mcssql,chess}—struggle when applied on their own to this benchmark, revealing fundamental limitations in their ability to generalize and \textit{scale}. These shortcomings have led to growing interest in agentic Text-to-SQL methods~\cite{reforce,linkalign,agenticdata}, where agents iteratively interact with databases to inspect schemas, validate partial queries, and adapt to dialectal differences.

While agentic systems demonstrate improved adaptability to large and complex schemas, there is a challenging tradeoff between computational cost and accuracy that makes these systems difficult to scale. Agents often incur substantial \textbf{redundancy}, repeatedly retracing identical exploration steps across queries on the same database, which inflates execution cost and latency. They are also prone to \textbf{planning variance}, where suboptimal initial reasoning paths lead to inconsistent or failed outcomes. Moreover, the inherently iterative nature of these systems results in \textbf{high computational cost and latency}, as each step consumes additional tokens and time. Under practical constraints such as step limits, token limits, or latency budgets, these inefficiencies significantly reduce both overall accuracy and efficiency~\cite{liu2025supportingaioverlordsredesigning}.

To address these limitations, we present Agent Semantic Memory (\sys{}), a scalable, stable, and efficient agentic framework for Text-to-SQL. Instead of treating each query as an isolated interaction, \sys{} introduces a structured semantic memory that captures and reuses trajectories from prior executions. Each trajectory is enriched and stored with semantic annotations to support future retrieval and reasoning. When a new query arrives on a known database, \sys{} reuses relevant portions of prior trajectories, eliminating redundant exploration and ensuring more consistent behavior. In addition, frequently co-occurring tool sequences are automatically coupled into composite tools, shortening trajectories and improving execution efficiency.

Beyond Text-to-SQL, \sys{} provides a generalizable foundation for other agentic data tasks—including information extraction, data cleaning, and data transformation—where structured trajectory reuse is critical for scalable and high-quality performance.

In summary, our contributions are as follows:

\begin{enumerate}
    \item We present \sys{}, an agentic framework that captures and  leverages structured semantic memory for enterprise-level Text-to-SQL.
    
    \item We design a structured semantic memory that encodes prior trajectories in an interpretable and retrievable format. This memory enables agents to retrieve relevant past experiences, improving efficiency and reasoning consistency.
    
    \item We introduce composite tools to \sys{} that streamline decision-making, reduce latency, and alleviate hallucination in agent planning and tool usage during complex multi-step query generation. These tools reduce both agent turns and token usage.
    
    \item We conduct extensive experiments demonstrate that \sys{} achieves state-of-the-art accuracy of 44.8\% on the Spider 2.0 Lite benchmark. Ablation studies further show that \sys{} effectively shortens average trajectory length by 25\% and improves execution accuracy by 35\%.
\end{enumerate}

\section{Background and Problem Formulation}
\label{sec:background}

\subsection{Agentic Text-to-SQL}
\label{sec:background:text2sql}

Modern ReAct-based agents~\cite{react} follow a standard pattern of reasoning and action steps. At each step, the agent observes the current state of the task, reasons about the next appropriate action, and executes an action that often involves tool use. For example, a code agent observes intermediate program states or execution outputs, reasons about modifications or next steps, and acts by generating and executing code to advance toward the final solution.

\begin{figure}[t]
  \centering
  \includegraphics[width=\columnwidth]{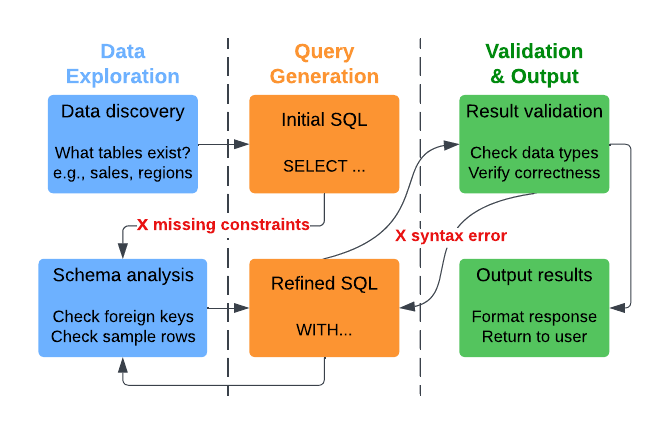}
  \vspace{-10pt}
  \caption{The standard workflow for Text-to-SQL agents is a series of steps alternating between data exploration, query generation, and answer validation.}
  \label{fig:example_workflow}
  \vspace{-10pt}
\end{figure}

Figure~\ref{fig:example_workflow} shows the standard workflow of an agentic Text-to-SQL system, which generally follows three phases: data exploration, SQL query generation/execution, and response validation. During the data exploration phase, the agent explores and understands the database schema, identifying relevant tables, columns, and values needed for the task. In enterprise environments, schema information is often parsed and stored offline in files, which the agent can reference alongside simple exploratory SQL queries to understand the data. The agent then attempts to generate SQL queries to answer the given task, often in parts, before synthesizing a final query. Without encountering any errors, the agent can choose to validate its response or directly output it. When errors occur, such as syntax violations or unexpected query results, the agent can revise the SQL or return to data exploration for additional context, depending on the nature of the error.

We identify three key opportunities within these standard workflows that motivate our proposed solution.

\subsubsection{Repeated exploration}
Analysis of agent reasoning patterns reveals that data exploration is \textit{inherently repetitive} across queries on the same database. In fact, Liu et al.~\cite{liu2025supportingaioverlordsredesigning} highlight the repetitive nature of agent trajectories on the BIRD benchmark~\cite{li2024can}, reporting that fewer than 10–20\% of the trajectories are distinct.

We observe the same behavior for agents operating in general Text-to-SQL settings. In the initial steps, the agent consistently performs a set of basic exploration actions. It begins by reading schema files or performing PRAGMA queries to obtain understanding of the database schema, including table structures, column names, and data types. If external knowledge files are available, the agent reads them for additional context about the database or task. If the agent requires additional help dealing with particularly large schemas, then it consults the help of tools, such as vector search, or issues additional SQL queries to understand the contents of candidate tables.

\heading{Opportunity 1} 
\textit{Reuse prior trajectories} to eliminate redundant exploration and improve efficiency on new questions over the same database.

\subsubsection{Strategy selection}
Traditional Text-to-SQL systems have converged toward highly precise, pipeline-style solutions that consist of a fixed sequence of steps to every question~\cite{chasesql,chess}. Similarly, agentic Text-to-SQL approaches remain heavily reliant on prompt engineering~\cite{reforce}, where carefully designed instructions define the agent's workflow and tool use to ensure consistency. 

\begin{figure}[t]
  \includegraphics[width=\columnwidth]{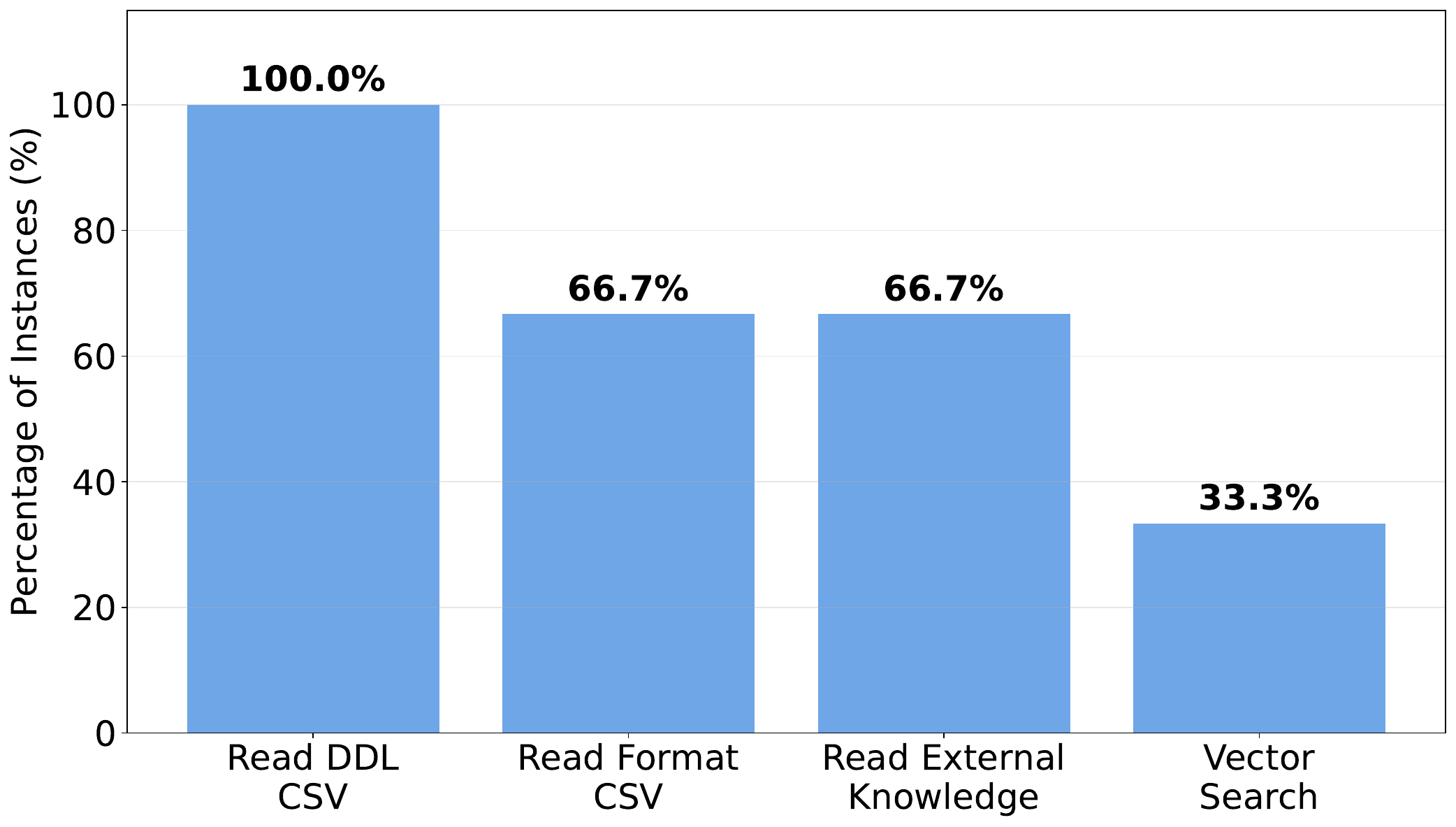}
  \vspace{-15pt}
  \caption{Distribution of the first seven steps in agent trajectories on the firebase from Spider 2~\cite{lei2024spider}, reporting the percentage of instances in which each specific action is taken.}
  \label{fig:example_firebase}
  \vspace{-10pt}
\end{figure}

However, enforcing a single dominant strategy across all queries can be suboptimal in terms of both accuracy and efficiency. While standardized routines can be effective for common cases, they often fail on out-of-distribution queries. For example, Figure~\ref{fig:example_firebase} reveals that the agent chose to perform vector search in only 30\% of questions on the `firebase' database in Spider 2.0, likely due to the nested schema and sufficient prior knowledge gained from prior exploration steps. In such cases, adhering to a rigid strategy can introduce irrelevant context, unnecessary computation, and wasted steps that otherwise contribute to more targeted actions to solve the given query.

\heading{Opportunity 2}
\textit{Dynamically adapt trajectories} according to the characteristics of each task and database, rather than relying on a fixed, one-size-fits-all strategy.

\subsubsection{Reducing variance}
Agent behavior in Text-to-SQL generation inherently exhibits variance across runs. Small deviations, including trivial syntax errors, in intermediate steps can derail the reasoning process, causing the agent to abandon a correct reasoning path and pursue an alternative one. This instability leads to inconsistent query generation and, consequently, variation in final results~\cite{wu2025agenticreasoningstreamlinedframework}.

Deterministic behavior cannot be enforced reliably even by setting the model temperature to zero for agents. Moreover, even if a strictly deterministic setting was possible, it prevents the agent from exploring alternative reasoning paths, limiting its ability to discover solutions to previously unseen or complex tasks~\cite{zhao2025llmbasedagenticreasoningframeworks}. Common Text-to-SQL strategies, such as candidate generation and query refinement, explicitly rely on controlled model variance to improve the likelihood of a correct SQL being generated. However, once new contexts are introduced into the reasoning workflow, subsequent steps can diverge unpredictably, even under deterministic settings.

\vspace{-5pt}
\heading{Opportunity 3}
Optimize the \textit{trade-off between tool complexity and trajectory length} to minimize variance and enhance accuracy and robustness.

\subsection{Problem Statement}
\label{sec:background:ps}


In this paper, we aim to to develop an agentic Text-to-SQL system that, given a natural language query $q$, a database $D$ and a set of tools $\mathcal{U}$, produces an optimized reasoning trajectory
\[ \tau(q, D, u) = \text{arg} \max_{\tau \in T(q, D, \mathcal{U})} \text{Acc}(\tau) \]
where $T(q,D, u)$ denotes the space of executable reasoning trajectories composed of tool usages and intermediate reasoning steps, and Acc($\tau$) measures the accuracy of the generated SQL query. 

\section{Methodology}
\label{sec:methodology}

\begin{figure*}[t]
    \centering
    \includegraphics[width=\linewidth]{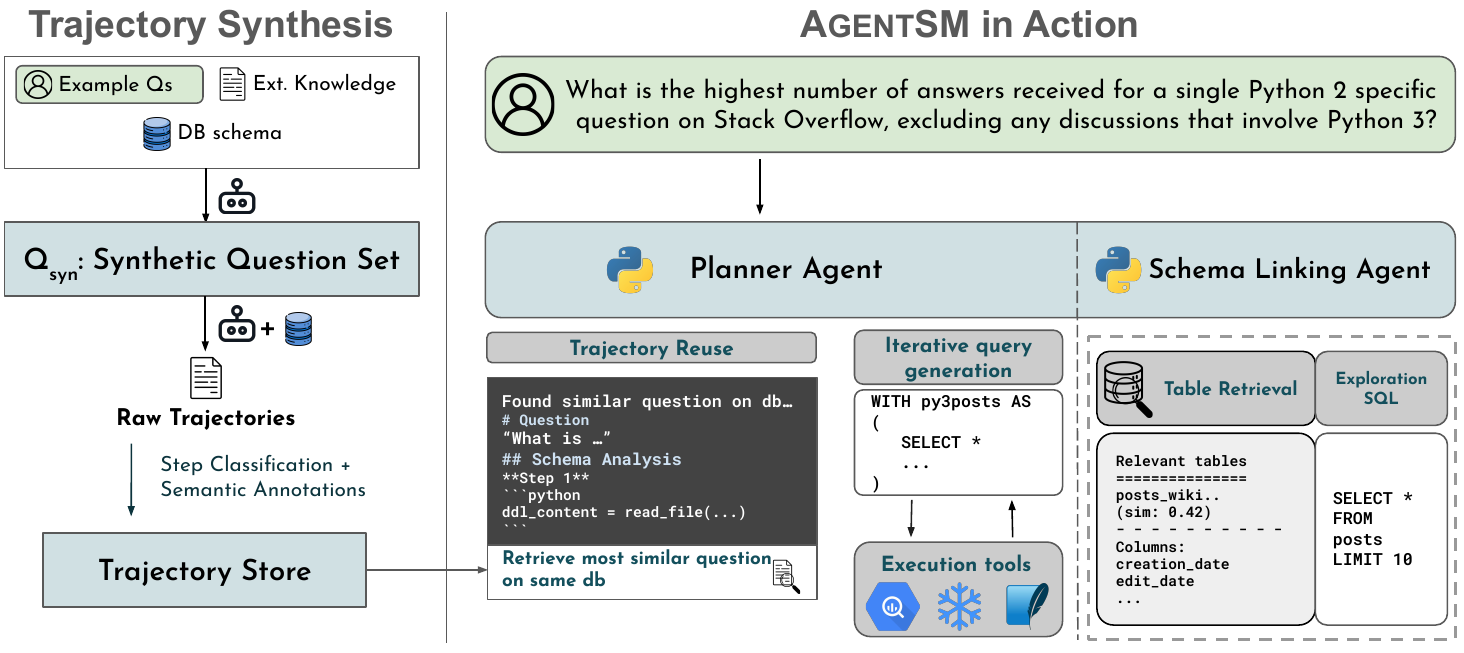}
    \caption{An overview of the \sys{} architecture that leverages trajectories in structured semantic memory.}
    \label{fig:architectureoverview}
\end{figure*}

Figure~\ref{fig:architectureoverview} depicts a full overview of the \sys{} framework, which consists of two agents with a trajectory-reading method and careful tool design to improve agent performance on Text-to-SQL tasks.

\subsection{\sys{} Architecture}
\label{sec:methodology:arch}
\vspace{-10pt}

\heading{Agents} We introduce two main agents in our framework, the planner agent and the schema linking agent. Inspired by the ReAct framework, agents alternate between observation, reasoning, and action cycles. In our implementation, both agents are coding agents that take actions by writing Python code. Unlike prior methods, however~\cite{lei2024spider}, we do not restrict the functions or code that the agent can write to a limited, predefined action space. Instead, we provide a broad range of authorized libraries which the agent may use to help with data exploration and analysis, including Pandas and NumPy.

The \textbf{planner agent} serves as the system's core reasoning component. It constructs a high-level execution plan, iteratively issues SQL queries, performs reasoning and data wrangling, and applies query refinements before producing the final answer. Unlike multi-agent pipelines that delegate SQL generation to a dedicated agent\cite{wang2025macsqlmultiagentcollaborativeframework}, we intentionally integrate SQL generation within the planner’s reasoning loop. This design choice ensures that the planner agent directly leverages the schema and data understanding gathered through exploration, avoiding the context loss commonly observed when control is handed off to specialized sub-agents. Moreover, by letting the planner generate and execute SQL, \sys{} avoids unnecessary inter-agent communication overhead and latency.

The \textbf{schema linking agent} is managed by the planner agent, and its role is to perform fine-grained data exploration for the task, when the planner agent requires deeper inspection of candidate tables. It has access to specialized tools including a vector search tool, which performs a similarity search on a precomputed index of schema information to narrow a list of tables and columns relevant for a given question. The schema linking agent operates within a small budget (e.g., 5 steps) to perform basic queries to interpret nested structures, probe candidate columns, and validate potential join paths, before returning its findings (i.e., mapping question tokens to schema elements) to the planner agent. Delegating deeper exploration to a separate agent allows the planner agent to not drift in its reasoning and keep only relevant information in context.

Although prior work has proposed complex multi-agent systems (e.g., introducing query fixers, decomposers, or evaluators~\cite{reforce}), we deliberately keep \sys{}'s architecture simple with two tightly coupled agents. This decision is guided by empirical findings~\cite{tran2025,gridach2025} that excessive agent decomposition leads to communication overhead, fragmented memory, and increased behavioral variance. In Text-to-SQL tasks, contextual knowledge is often highly localized: understanding a schema’s nested hierarchy requires maintaining persistent state over multiple exploration and reasoning cycles. Therefore, introducing additional agents would necessitate a shared memory~\cite{NEURIPS2024_fa54b0ed,rezazadeh2025} to prevent context fragmentation. However, shared memory introduces nontrivial challenges in retrieval efficiency, consistency management, and error propagation, which we leave as future work.

\heading{Tools} 
Each agent in \sys{} is equipped with a carefully selected set of tools designed to balance flexibility, robustness, and efficiency. The planner agent is provided with tools for trajectory retrieval, output validation, and result saving. The trajectory retrieval tool allows the planner to review reasoning traces from semantically similar questions before starting a new task, helping it avoid redundant exploration steps. The validation and saving tools ensure that the final SQL and results follow consistent formatting and can be correctly persisted. The schema linking agent, on the other hand, is equipped with tools specialized for fine-grained schema exploration, particularly the vector search tool that retrieves semantically relevant tables and columns from a precomputed index. Restricting this capability to the schema linking agent prevents redundant searches between the two agents.

Both agents share a set of general-purpose tools for local directory exploration and SQL execution on supported backends such as BigQuery, Snowflake, and SQLite. To improve execution robustness, we incorporate self-refinement into the SQL execution tools. When a query fails or returns an empty result, the SQL execution tools automatically attempt a correction based on the received feedback. This mechanism reduces the likelihood that a single execution error derails the reasoning trajectory.

In addition to these basic tools, we introduce \textit{composite tools}, special types of tool that combine the logic of single-purpose tools that frequently co-occur in consecutive steps. These composite tools, described in Section~\ref{sec:compositetools}, simplify planning, reduce unnecessary tool usage, and improve consistency across runs.

\subsection{Trajectory Synthesis and Retrieval}
\label{sec:methodology:trajectory}

\heading{Synthetic question exploration}
We make a key insight that data exploration is inherently reusable: once an agent has examined the schema, inspected columns, and issued exploratory queries, those interaction traces remain valuable for subsequent questions on the same database. Crucially, this property extends to \emph{synthetic questions}, questions that can be formulated and executed offline that are deliberately designed to elicit data exploration. By enriching the trajectory store with such exploration-rich traces, \sys{} effectively reduces redundant probing and enhances agent performance on real queries.

\setlength{\intextsep}{10pt} 
\setlength{\textfloatsep}{10pt}
\begin{algorithm}[t]
\small
\caption{Synthetic Question Generation. \\
GenQ($S,K,Q$) generates a new question using an LLM over the database schema $S$, external knowledge $K$, and existing questions $Q$ for $S\in\mathcal{D}$.\\
Allocation($S,n,p$) returns an integer budget $k$ for the max number of questions on $S$ based on the query distribution $p(\text{Q})$ and the remaining global budget $n$.}
\label{alg:synthetic_questions}
\KwIn{Databases $\mathcal{D}$, target count $n$, query distribution $p(\text{Q})$}
\KwOut{Synthetic question set $Q_{\text{syn}}$}

$Q_{\text{syn}} \gets \emptyset$\;

\ForEach{schema $S \in \mathcal{D}$}{ 
    $q \gets \text{GenQ}\big(S,\ K,\ Q\big)$\;
    $Q_{\text{syn}} \gets Q_{\text{syn}} \cup \{q\}$\;
}

\While{$|Q_{\text{syn}}| < n$}{
    $k \gets \text{Allocation}\big(S,\ n,\ p(\text{Q})\big)$\;

    \For{$i \gets 1$ \KwTo $k$}{
        \If{$|Q_{\text{syn}}| = n$}{\textbf{break}}
        $q \gets \text{GenQ}\big(S,\ K,\ Q\big)$\;
        $Q_{\text{syn}} \gets Q_{\text{syn}} \cup \{q\}$\;
    }
}
\Return $Q_{\text{syn}}$\;
\end{algorithm}

Algorithm~\ref{alg:synthetic_questions} outlines the process for generating the full synthetic question set $Q_{\text{syn}}$. First, we ensure that each database present in our query distribution is covered by at least one synthetic question (Lines 2-4). We allocate additional synthetic questions across databases proportionally to the distribution these databases on the query set, up to a fixed total question budget $n$ (Line 6). This allocation strategy naturally emphasizes large, complex schemas, where additional exploration traces are most beneficial (Lines 7-11).

For each selected schema $S$, we prompt an LLM with the schema, external knowledge file, and a set of existing questions to generate diverse candidate questions that involves different operators, tables, and columns. The number of questions generated per schema follows the same distribution as the full query set. The prompts used to generate candidate questions can be found in the repository\footnote{\url{https://tinyurl.com/xcey6h33}}.

Each synthetic question $q \in Q_{\text{syn}}$ is then consumed by an agent equipped with SQL and file-reading tools only to produce a trajectory $T(q)$, which is stored for reuse. Although the answers to these synthetic questions are not evaluated, the resulting trajectories provide dense and diverse exploration traces. When the agent encounters new queries in future, these traces provide rich context, allowing it to leverage past exploration rather than repeating it.

\heading{Step classification}
Agent workflows are non-linear, consisting of sequence of distinct states that the agent enters and exits during execution. In practice, a Text-to-SQL agent alternates among three primary phases depicted in Figure~\ref{fig:example_workflow}: (1) exploring the database to understand the data, (2) formulating and executing candidate queries, and (3) validating results or producing an answer. Similar stage decompositions have been observed in prior work~\cite{liu2025supportingaioverlordsredesigning}.

Grounding trajectory reuse in these phases is essential: exploration steps are often generalizable across questions, execution steps encode database-specific reasoning, and validation steps highlight patterns of error correction and output formatting. Step classification helps avoid presenting the agent with a noisy sequence of mixed tool invocations and rendering reuse ineffective.

To classify steps, we apply a lightweight text-pattern matching using regular expressions rather than relying on step position in the trajectory. While exploration often occurs early, agents may re-enter this exploration phase later when encountering execution errors or missing context. We therefore identify intent based on tool usage and query patterns. For example, file listing and reading operations are matched as exploration, while complex SQL queries beginning with `\texttt{WITH}` (common table expressions) are almost always recognized as main query execution. We also explored LLM-based step classification and observed comparable accuracy but substantially higher cost and latency.

Each classified step is parsed and saved in a separate trajectory file. During trajectory reading, the agent selectively loads only the trajectory segment corresponding to the relevant phase for reference. For example, during data exploration phase, the agent retrieves and reads the relevant exploration trajectories that include basic schema analysis and external knowledge reading actions, enabling efficient reuse of prior context without redundant steps.

\heading{Trajectory structure}
Agent trajectories are typically long, complex sequences that interleave text, code, reasoning traces, and execution outputs. Reading raw trajectories introduces significant noise and increases the \textit{lost-in-the-middle} effect, where valuable information is buried amid verbose agent debugging logs and intermediate outputs. 

\begin{table}[t!]
\centering
\begin{tabular}{c|c|c}
\toprule
\textbf{Method} & \textbf{Avg Steps} & \textbf{Accuracy (\%)} \\
\midrule
No trajectory & 22.62 & 25 \\
Naive         & 22.62 & 25 \\
Markdown      & 16.50 & 50 \\
JSON          & 15.12 & 50 \\
\bottomrule
\end{tabular}
\caption{Average trajectory length (in steps) and execution accuracy on a sample of Spider~2.0~Lite questions, under different trajectory formats. Reading raw, unstructured trajectories yield no improvement compared to without using any trajectory. In contrast, structured representations, in either Markdown or JSON format, significantly reduce the average number of steps and improve the overall execution accuracy.}
\label{tab:traj_formats}
\end{table}

To help the agent extract useful knowledge more effectively, we impose a structured representation on each trajectory. In particular, we store agent trajectories in markdown format, with semantically meaningful headers automatically generated by a lightweight LLM (e.g., Claude Haiku 4.5). This structure segments the trajectory into interpretable sections, improving both retrieval and readability.

As shown in Table~\ref{tab:traj_formats}, structured trajectories, whether represented in markdown or JSON, yield substantial improvements in both execution accuracy and average step reduction compared to unstructured logs, based on a sample of 20 benchmark questions. We adopt markdown as the format for its consistency with external knowledge files and human readability. The implementation details of step classification and trajectory structure generation can be found in the repository\footnote{\url{https://tinyurl.com/492ymhcn}}.


\heading{Trajectory selection}
Having established a structured representation for storing agent trajectories, we next describe how these stored trajectories are retrieved and reused during inference. Among the synthetic questions $Q_{syn}$ for which trajectories are synthesized and saved, we first restrict retrieval to those associated with the same database as the given question $q$. This filtering narrows the search space to relevant trajectories:
\begin{equation}
Q_d = \{q' \in Q_{\text{syn}} \mid \text{db}(q') = \text{db}(q)\}
\end{equation}

From this filtered subset $Q_d$, we then select the most relevant trajectory $T$ based on its associated question $q'$ that has the highest semantic similarity to the given question $q$:

\begin{equation}
q^* = \arg\max_{q' \in Q_d} \text{sim}(q', q)
\end{equation}

\begin{equation}
T = \text{traj}(q^*)
\end{equation}

This strategy is effective for data exploration reuse, as semantically similar questions tend to probe the same set of tables during their initial exploration steps. The agent naturally prioritizes tables with their schema semantically similar to the question before performing deeper searches. Consequently, selecting trajectories based on question similarity leads to effective trajectory reuse.
Further improvements in retrieval quality could be achieved through finer-grained alignment between questions and stored trajectories. For example, schema-level retrieval could identify which tables are relevant to a question and restrict trajectory retrieval to those involving the same tables. Alternatively, trajectory-level alignment could leverage the agent’s initial plan to retrieve similar prior executions and adjust the reasoning strategy before execution. We leave these directions to future work, as fine-grained retrieval risks missing relevant context essential for accurate reasoning.


\begin{figure}[t]
\centering
\subfloat[Dist. of tool calls per step over 327 steps]{
    \includegraphics[width=0.96\columnwidth]{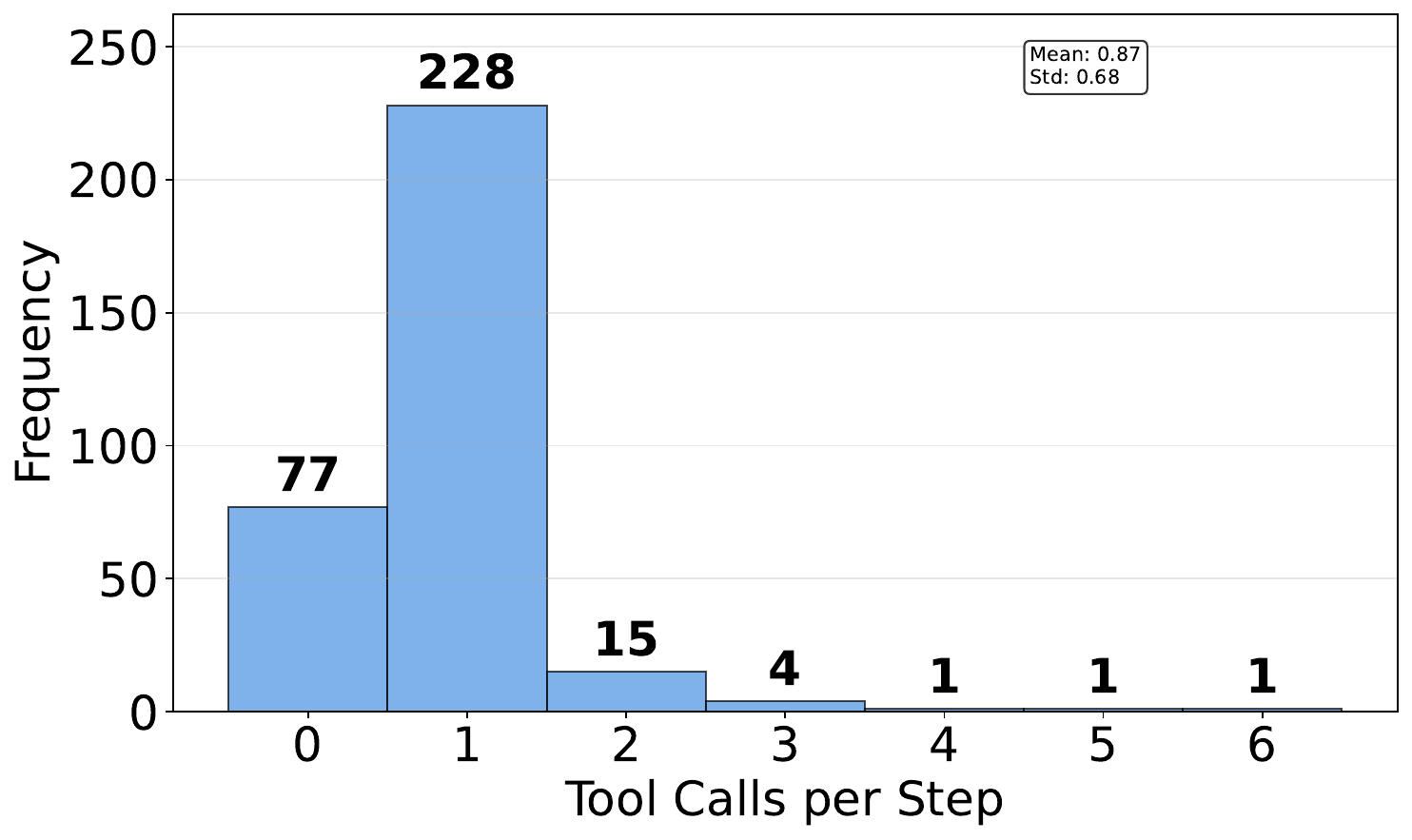}
    \label{fig:toolcallsperstep}
}\\
\subfloat[Composite tool formation.]{
    \includegraphics[width=0.8\columnwidth]{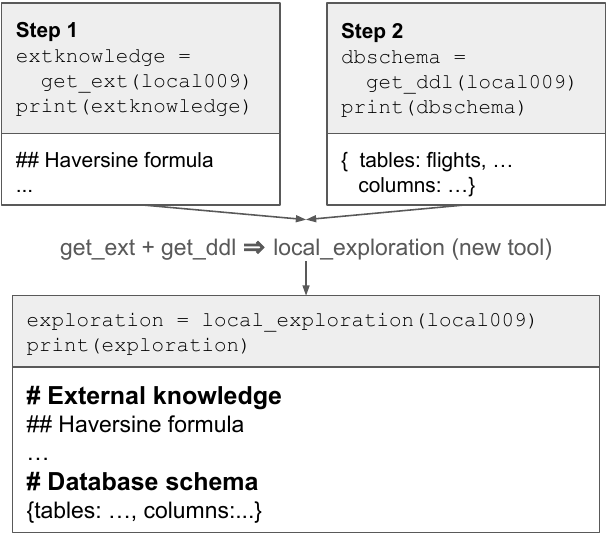}
    \label{fig:compositetoolexample}
}
\vspace{-6pt}
\caption{Analysis of tool usage and composite tool construction. (a) shows that most steps use zero or one tool, motivating composition. (b) illustrates how frequently co-occurring tools are merged into a higher-level composite tool.}
\label{fig:toolanalysis}
\end{figure}

\subsection{Composite Tools}
\label{sec:compositetools}

\heading{Composite tool construction}
Across trajectories, we observe that certain tool pairs consistently co-occur in sequence. Figure~\ref{fig:compositetoolexample} illustrates how the \texttt{get\_ext} $\rightarrow$ \texttt{get\_ddl} pattern frequently appears when the agent needs to access an external knowledge file and the database schema at the beginning of the exploration phase. As such, these tools can be combined into a single composite tool \texttt{local\_exploration}.

As shown in Figure~\ref{fig:toolcallsperstep}, agents rarely chain multiple tool invocations within a single reasoning step, instead issuing one call per step, even when such sequences are highly regular and deterministic. This pattern leads to inefficiency: the agent spends valuable steps on executing repetitive exploration routines rather than focusing on higher-level reasoning and problem solving. 

To mitigate this overhead, we introduce \emph{composite tools}, which combine commonly co-occurring tool sequences into single higher-level operations. Formally, we construct a composite tool:
\[
C = \langle t_1, t_2, \dots, t_k \rangle \quad \text{if} \quad 
\frac{f(t_1, t_2, \dots, t_k)}{N} \geq \tau
\]
where $t_i \in \mathcal{T}$, $\mathcal{T}$ is the tool set, $f(\cdot)$ counts the trajectories containing the subsequence $(t_1, t_2, \dots, t_k)$, $N$ is the total number of trajectories, and $\tau$ is a chosen support threshold.

In practice, \sys{} follows several heuristics to construct effective composite tools while preserving modularity. First, tools are combined only when they co-occur within the same reasoning phase (e.g., data exploration or validation), ensuring that composite tools remain semantically coherent. Second, tools that appear across multiple phases are excluded from composition to preserve their agility for reuse in different contexts. Finally, to avoid excessive aggregation, we constrain the maximum size of composite tools, preventing all tools collapse into a single monolithic operator.

Each composite tool is then assigned a descriptive name and natural language summary by an LLM, which the agent can reference during prompting. At inference time, the agent uses these composite tools directly, effectively replacing repeated multi-step exploration with single high-level actions and freeing more trajectory budget for reasoning.


\section{Evaluation}
\label{sec:eval}

In this section, we evaluate the effectiveness of \sys{} on Spider 2, a widely recognized benchmark for agentic Text-to-SQL systems. Our evaluation is designed to answer the following key questions:

\begin{enumerate}
    \item How do trajectory synthesis and retrieval influence the agent’s efficiency and accuracy?
    \item How does the optimized tool design improve agent's efficiency and accuracy?
\end{enumerate}

\subsection{Evaluation Setup}
\label{sec:eval:setup}
\vspace{-10pt}

\heading{Dataset} Spider 2.0~\cite{lei2024spider} is a large-scale benchmark of 547 Text-to-SQL workflow problems that emulate the challenges of enterprise-scale data analysis. Unlike traditional Text-to-SQL benchmarks, Spider 2.0 requires solving tasks that involve extremely long contexts, nested schemas, and diverse SQL dialects. The databases in Spider 2.0 often span thousands of columns and are deployed in production systems such as BigQuery and Snowflake.

This makes Spider 2.0 a natural workload for agentic approaches, where agents must iteratively plan and issue queries to reach a final answer. We select this benchmark to evaluate our method, as our framework is specifically designed to operate in such complex environments that reward iterative reasoning and tool use.

\begin{table*}[ht!]
\centering
\begin{tabular}{l|cccc|ccc|c}
\hline
\multirow{2}{*}{Method} & \multicolumn{4}{c|}{EX (\%)} & \multicolumn{3}{c|}{Avg. Length} & \multirow{2}{*}{Avg Latency (s)} \\
\cline{2-5} \cline{6-8}
 & BigQuery & Snowflake & SQLite & Overall & Steps & Input Toks & Output Toks &  \\
\hline
SpiderAgent (claude-3-7-sonnet) & 37.6 & 12.6 & 40.7 & 28.7 & 18.9 & 200K & 4K & 363.2 \\
SpiderAgent (claude-4-sonnet)   & -    & -    & -    & 27.8$^*$ & - & - & - & - \\
CodingAgent (claude-4-sonnet)   & 27.3 & 11.1 & 42.2 & 24.7 & 18.1 & 200K & 4K & 325.5 \\
\hline
AgentSM (claude-3-7-sonnet) & 40.5 & 33.3 & 42.2 & 38.4 & 16.8    & 299K    &  5K  & 226.4 \\
AgentSM (claude-4-sonnet)   & \textbf{52.2} & \textbf{35.0} & \textbf{51.9} & \textbf{44.8} & 16.4 & 300K & 5K  & 247.1 \\
\hline
\textit{AgentSM (claude-4-sonnet, gold tables)} & 62.0 & 44.0 & 79.2 & 57.6 & 15.8 & 268K & 5K & 194.3 \\
\hline
\end{tabular}
\caption{Comparison of execution accuracy, step/token length, and latency for methods on the Spider 2.0 Lite dataset. $*$This result is quoted directly from the original SpiderAgent paper.}
\label{tab:mainresults}
\vspace{-15pt}
\end{table*}

\heading{Metrics} We measure execution accuracy as the percentage of execution output matches to the gold output using the evaluation function provided by the authors of the original benchmark. We adopt this metric as per the original benchmark. We report accuracy for BigQuery, Snowflake, and SQLite questions, along with an overall result.

We also report the average agent trajectory length in both reasoning steps, as well as the number of input and output tokens. We also report average end-to-end latency for a single execution. 

\heading{Experimental setup} We evaluate our method using the Claude 3-7~\cite{anthropic2025claude37sonnet} and Claude 4 Sonnet~\cite{anthropic2025claude4sonnet} models. Several agent frameworks have risen in popularity, including LangChain~\cite{langchain2025}, smolagents~\cite{smolagents2025}, etc. While our approach is framework-agnostic, for evaluation we implement it using smolagents as the foundation of our agentic Text-to-SQL system.

For all evaluated systems, we employ the MiniLM-L6-v2 model~\cite{reimers2019sentencebert} from SentenceTransformers for generating embeddings and use the FAISS~\cite{johnson2019billion} library for vector similarity search. We run ArcticSQL-7B with vLLM~\cite{kwon2023efficient} on a cluster of 8$\times$A100 80GB GPUs.

\heading{Baselines} For our main experiments on Spider 2.0 Lite, we compare \sys{} against two baselines: SpiderAgent~\cite{lei2024spider} and a standard coding agent (CodingAgent) implemented using the smolagents~\cite{smolagents2025} framework equipped with basic SQL execution tools described in Section~\ref{sec:background}. We evaluate both baselines with Claude 3-7~\cite{anthropic2025claude37sonnet} and Claude 4 Sonnet~\cite{anthropic2025claude4sonnet} as the underlying LLMs. The performance of other state-of-the-art systems is available on the official Spider 2.0 leaderboard\footnote{\url{https://spider2-sql.github.io/}}.

\subsection{Main Results}
\label{sec:eval:main}

Table~\ref{tab:mainresults} contains the execution accuracy for each method on the full Spider 2.0 Lite benchmark with 547 questions. We report execution accuracy for BigQuery instances, Snowflake instances, and SQLite instances along with overall accuracy. 

\heading{Accuracy} \sys{} substantially outperforms the prior agentic baseline SpiderAgent~\cite{lei2024spider}, achieving higher accuracies across each SQL dialect and an overall 14.1\% improvement in execution accuracy. \sys{} also outperforms a standard coding agent implemented using the same agent framework\cite{smolagents2025} and model~\cite{anthropic2025claude4sonnet} by over 20\%. Based on our evaluation, \sys{} would rank No. 1 on the Spider 2.0 Lite leaderboard, with an overall accuracy of 44.8\% at the time of paper submission. Notably, we achieve this result without using a powerful reasoning model such as Qwen3~\cite{qwen3} or OpenAI's o3~\cite{openai-o3}, which other top systems rely on.

Across all SQL dialects, the questions on the Snowflake database are the most challenging for the agents. The platform has more questions on databases with larger and nested schemas. For this platform alone, our method performs better with an older model Claude 3-7 Sonnet than Claude-4 Sonnet.

\heading{Efficiency} To understand how \sys{} improves agent efficiency, we revisit the standard agentic workflow (i.e., CodingAgent) and analyze how our method reshapes the agent’s progression through three key stages: database exploration, query execution, and answer validation. We randomly sample 75 questions from the Spider~2.0~Lite dataset and classify each step in the agent's trajectory as one of the three stages, based on the reasoning trace and executed queries. We aggregate these stage annotations across instances to measure the proportion of each trajectory devoted to exploration versus execution.

\begin{figure}[!t]
  \centering
  \includegraphics[width=0.9\columnwidth]{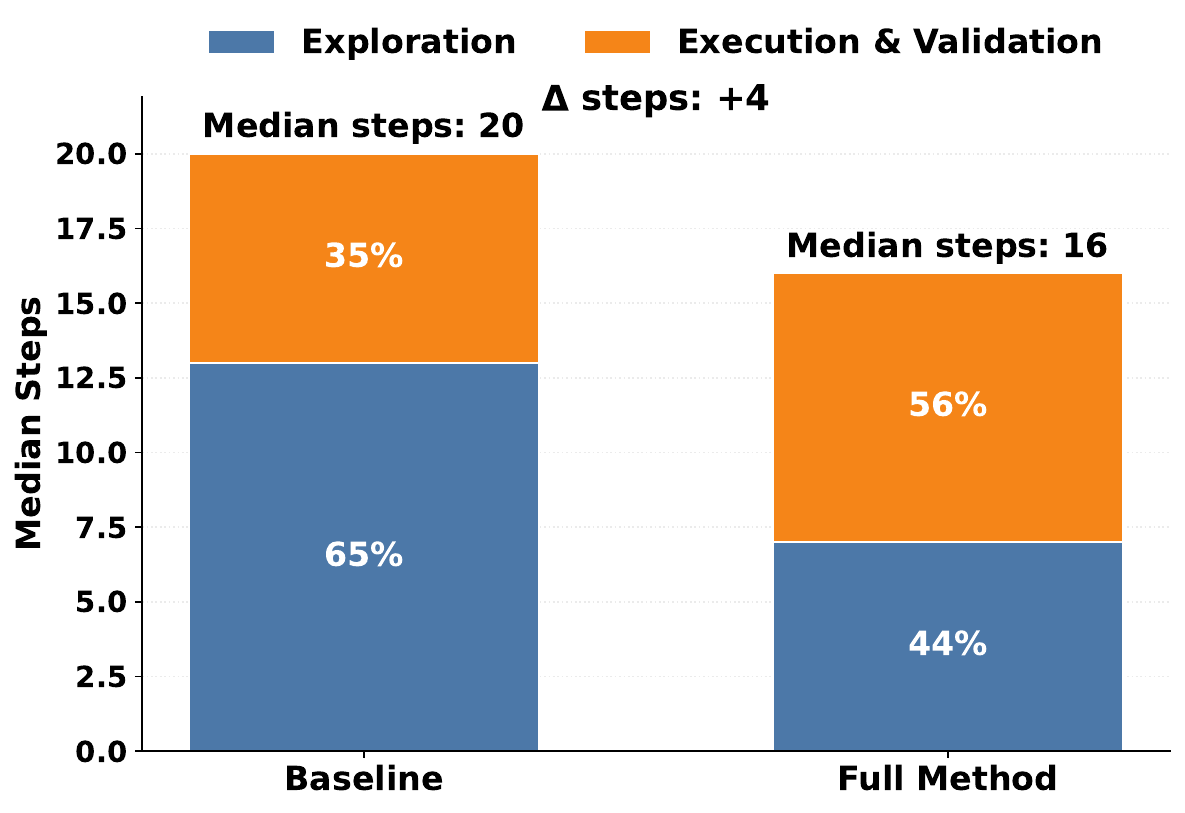}
  \caption{Trajectory composition: median steps split into exploration vs.\ execution and validation.}
  \label{fig:agentstage}
\end{figure}

As shown in Figure~\ref{fig:agentstage}, \sys{} transitions from exploration to execution significantly earlier, completing tasks with fewer overall steps. In contrast, a standard coding agent not utilizing synthesized trajectories repeatedly issues exploratory queries and spends most of its steps analyzing schemas rather than composing or validating SQL queries. By leveraging structured semantic memory, \sys{} alleviates redundant data exploration, maintains consistent reasoning efficiency, and scales gracefully to larger databases and more complex questions.

\subsection{Ablation Studies}
\label{sec:eval:ablation}

\subsubsection{Effectiveness of trajectory reading and composite tools}
We conduct ablation studies on two key components of \sys{}, the synthesized trajectory and the composite tools, to report their respective effects on accuracy and efficiency.

\begin{table*}[ht!]
\centering
\begin{tabular}{l|cc|ccc|ccc|cc}
\hline
\multirow{2}{*}{Method} 
 & \multicolumn{2}{c|}{Execution Accuracy} 
 & \multicolumn{3}{c|}{Steps} 
 & \multicolumn{3}{c|}{Avg. Length} 
 & \multicolumn{2}{c}{Latency} \\
\cline{2-11}
 & EX (\%) & $\Delta_{\text{EX}}$ 
 & Total & Avg & $\Delta_{\text{Steps}}$ 
 & Input Toks & Output Toks & $\Delta_{\text{Input}}$ 
 & Time (s) & $\Delta_{\text{Time}}$ \\
\hline
No trajectory reading & 17.3 & --34.7 & 1527 & 20.36 & +4.37 & 360K & 5.6K & +8K & 348.6 & +96.2 \\
No composite tools    & 16.2 & --35.8 & 1557 & 20.76 & +4.77 & 398K & 5.8K & +46K & 523.7 & +271.3 \\
\textbf{Full method}  & \textbf{52.0} & \textbf{--} & \textbf{1167} & \textbf{15.99} & \textbf{--} & \textbf{352K} & \textbf{5.2K} & \textbf{--} & \textbf{252.4} & \textbf{--} \\
\hline
\end{tabular}
\caption{Effects of synthesized trajectory and composite tools on execution accuracy, step count, token usage, and latency over 75 sampled questions. 
Deltas are computed relative to the full method. 
Removing either component increases token usage and latency while sharply decreasing execution accuracy.}
\label{tab:ablation}
\vspace{-15pt}
\end{table*}

Table~\ref{tab:ablation} reports the average trajectory length and accuracy for a sample of 75 examples from Spider 2.0 Lite with and without trajectory reading. We observe that enabling trajectory reading reduces the average trajectory length of the agent by 25\% while improving accuracy by 35\%. This is consistent with the results reported in Section~\ref{sec:eval:main}. 

Composite tools have a similarly pronounced impact on agent trajectory length and accuracy. Without composite tools, the agent requires the largest number of steps across all sample instances, yet achieving the lowest overall execution accuracy.

The observed accuracy gains naturally stem from the reduction in average trajectory length, which can be attributed to two factors: (1) longer trajectories often correspond to inherently more complex questions, and (2) as trajectory length increases, the agent becomes more likely to reach the maximum step threshold before solving the question. By reducing the number of steps spent on data exploration, \sys{} effectively mitigates the issues, allowing the agent to allocate more steps toward reasoning and to iteratively refine its final queries before reaching the maximum steps.

\subsubsection{Impact of gold tables}
We further evaluate \sys{} on Spider~2.0~Lite with gold tables (as provided by Spider 2.0). Using the gold tables and columns, our method achieves \textbf{55.3\%} execution accuracy on the question subset (listed in Table~\ref{tab:mainresults}). The results indicate a substantial improvement in execution accuracy compared to the base configuration of our method. This highlights that schema linking is still a significant hurdle in these complex database scenarios, as agents often struggle to identify the right tables and columns when data and the given question are ambiguous.


\vspace{-5pt}
\subsection{Error Analysis}
\label{sec:exp:eval_error}

We provide a detailed error analysis on the full Spider 2.0 Lite evaluation to better understand the remaining failure cases of \sys{}. Among the 245 incorrect instances, 34\% occur on BigQuery, 22\% on SQLite, and 44\% on Snowflake, which shows the highest relative error rate. Within Snowflake, 5\% of failures stem from exceeding the step budget, 30\% from schema-linking errors caused by nested schemas, and the remainder from logical or dialect-specific syntax errors during query generation.


We further examine how structured semantic memory influences the agent’s success rate.
When relevant trajectories are available, \sys{} can reuse past reasoning steps to provide useful context for new queries. This is particularly effective for schema-linking, where prior trajectories often highlight useful table and join patterns, helping the agent quickly identify the relevant portions of a large schema. In contrast, the benefit is limited for queries that demand complex mathematical operations or intricate CTE reasoning. These tasks typically require deeper reasoning or problem decomposition that cannot be easily inferred from previous trajectories.

Finally, we observe that \sys{}'s performance varies across domains. Commonly used databases (e.g., \texttt{city}, \texttt{weather}, \texttt{census}) achieve 60–78\% accuracy, whereas heterogeneous databases (e.g., \texttt{github\_repos}, \texttt{idc}) lag behind at 14–40\%. These results suggest that \sys{} excels in domains with consistent schema patterns and recurring reasoning structures, but remains challenged by highly domain-specialized data sources.

\vspace{-5pt}
\section{Related Work}
\label{sec:related_work}
\vspace{-10pt}

\heading{Text-to-SQL} 
Recent Text-to-SQL work has spanned supervised fine-tuning, prompt engineering, and reinforcement learning based solutions. Supervised systems such as DIN-SQL~\citep{pourreza2023dinsqldecomposedincontextlearning} and MAC-SQL~\citep{wang2025macsqlmultiagentcollaborativeframework} utilize a pipeline with structured decoding, schema linking, and decomposition components to generate SQL queries. Prompting based methods~\cite{reforce, chasesql} leverage LLMs in few-shot or zero-shot settings, often augmented with modular pipelines that include candidate generation, majority voting, and execution verification. More recently, reinforcement techniques using group relative policy optimization (GRPO) engineer rewards to improve model performance on Text-to-SQL tasks~\cite{cheng2024text2sqlrl}. Multi-agent systems, such as AgenticData, explore utilizing a combination of planning, exploration, and validation agents to iteratively work towards an answer~\citep{agenticdata}. These techniques have steadily improved overall performance on the Spider~\citep{yu2019spiderlargescalehumanlabeleddataset} and BIRD~\citep{li2024can} benchmarks, however they often assume clean, well-aligned schemas and struggle with ambiguity, nested queries, and relational complexity found in enterprise settings. 

\heading{Coding Agents} 
Recent work on coding agents explore how LLMs can use external tools, such as web-search and code, to solve tasks through iterative reasoning. Approaches like ReACT~\cite{yao2022react} and Reflexion~\cite{shinn2023reflexion} highlight the effectiveness of planning, execution, and feedback stages for complex tasks. In line with these approaches, agentic Text-to-SQL systems decompose query generation into steps of interleaved tool executions and reasoning, allowing models to inspect schemas, validate partial queries, and iteratively refine the final answer. SpiderAgent~\citep{lei2024spider} formalizes this trend as an initial solution for Spider2.0. Recent works~\cite{liu2024codeact} have also extended this broad strategy with memory-guided refinement. While effective, most of these solutions rely on a fresh state for the agent or maintain some intra-task memory, unlike our solution which leverages inter-task memory to re-use database exploration reasoning across examples.

\heading{Agent Memory} 
Research on agentic memory can be broadly categorized into action-oriented memory, focused on persisting agent state, and knowledge-oriented memory, focused on persisting the knowledge agents gain from interaction. The most basic approach treats the language model's context window as a scratchpad-like working memory, where agents maintain notes. MemGPT~\citep{packer2024memgptllmsoperatingsystems} addresses the challenge of limited context windows by implementing a memory hierarchy that mimics operating system memory management, effectively bridging the previous two approaches. More recently, Mem0\citep{mem0} utilized graph representations of memory, maintaining memory with standard graph knowledge base structure: a collection of nodes (entities) and edges (relationships between entities). These forms of memory are limited in the amount of structure they can represent in their information for the agent.

\vspace{-10pt}
\section{Conclusion}
\label{sec:conclusion}

In this paper, we introduce \sys{}, a framework that enables agents to reuse reasoning steps across related Text-to-SQL tasks within the same database. By exploiting the inherent repetition in data exploration, \sys{} synthesizes, stores and retrieves structured trajectories to substantially reduce redundant exploration and improve execution efficiency. Moreover, our solution, equipped with semantic memory, enhances scalability, allowing agents to handle larger schemas and more complex queries. As a result, \sys{} delivers significant gains in efficiency across agent turns, token usage, and latency. On the Spider 2.0 Lite benchmark, \sys{} achieves an execution accuracy of 44.8\%.



\bibliographystyle{ACM-Reference-Format}
\bibliography{iclr2026_conference}

\end{document}